%% file: main.tex
\documentclass{article}

\usepackage[preprint,nonatbib]{neurips}

\usepackage[utf8]{inputenc} 
\usepackage[T1]{fontenc}    
\usepackage{hyperref}       
\usepackage{url}            
\usepackage{booktabs}       
\usepackage{amsfonts}       
\usepackage{nicefrac}       
\usepackage{microtype}      
\usepackage{xcolor}         
\usepackage{ulem}

\usepackage[numbers]{natbib}
\usepackage{paralist}
\usepackage{amsmath} 
\usepackage{adjustbox,booktabs,multirow}
\usepackage{multicol}
\usepackage{arydshln}
\usepackage{mdframed} 
\usepackage{wrapfig}
\usepackage{diagbox}
\usepackage{cleveref}
\usepackage{xspace}

\crefname{equation}{Eq.}{Eqs.}
\crefname{table}{Table}{Tables}
\crefname{figure}{Figure}{Figures}
\crefname{section}{Section}{Sections}
\crefname{algorithm}{Algorithm}{Algorithms}

\newcommand{\tg}{{T$_g$}\xspace}
\newcommand{\ffv}{{FFV}\xspace}
\newcommand{\tc}{{TC}\xspace}
\newcommand{\rg}{{R$_g$}\xspace}
\newcommand{\density}{{Density}\xspace}

\title{Open Polymer Challenge: Post-Competition Report}

\author{
    \normalfont Gang Liu$^{1}$ \quad
    Sobin Alosious$^{1}$ \quad
    Subhamoy Mahajan$^{2}$ \quad
    Eric Inae$^{1}$ \quad
    Yihan Zhu$^{1}$ \quad
    Yuhan Liu$^{1}$ \\
    Renzheng Zhang$^{1}$ \quad
    Jiaxin Xu$^{1}$ \quad
    Addison Howard$^{3}$ \quad
    Ying Li$^{2}$ \quad
    Tengfei Luo$^{1}$ \quad
    Meng Jiang$^{1}$ \\
    \\
    $^{1}$University of Notre Dame \quad
    $^{2}$University of Wisconsin–Madison \quad
    $^{3}$Kaggle \\
    \\
    \textit{Corresponding emails: {\tt \{gliu7, mjiang2\}@nd.edu}}
}

\begin{document}

\maketitle

\begin{abstract}
Machine learning (ML) offers a powerful path toward discovering sustainable polymer materials, but progress has been limited by the lack of large, high-quality, and openly accessible polymer datasets. The Open Polymer Challenge (OPC) addresses this gap by releasing the first community-developed benchmark for polymer informatics, featuring a dataset with 10K polymers and 5 properties: thermal conductivity, radius of gyration, density, fractional free volume, and glass transition temperature. The challenge centers on multi-task polymer property prediction, a core step in virtual screening pipelines for materials discovery. Participants developed models under realistic constraints that include small data, label imbalance, and heterogeneous simulation sources, using techniques such as feature-based augmentation, transfer learning, self-supervised pretraining, and targeted ensemble strategies. The competition also revealed important lessons about data preparation, distribution shifts, and cross-group simulation consistency, informing best practices for future large-scale polymer datasets. The resulting models, analysis, and released data create a new foundation for molecular AI in polymer science and are expected to accelerate the development of sustainable and energy-efficient materials. Along with the competition, we release the test dataset at \url{https://www.kaggle.com/datasets/alexliu99/neurips-open-polymer-prediction-2025-test-data}. We also release the data generation pipeline at \url{https://github.com/sobinalosious/ADEPT}, which simulates more than 25 properties, including thermal conductivity, radius of gyration, and density.

\end{abstract}

\section{Introduction}\label{sec:intro}
\input{1intro}

\section{Competition Overview}\label{sec:overview}
\input{2overview}

\section{Data Generation Pipeline}\label{sec:pipeline}
\input{3pipeline}

\section{Competition Results and Discussion}\label{sec:result}
\input{4result}

\section{Conclusion}\label{sec:conclusion}
\input{5conclusion}

\bibliographystyle{plainnat}
\bibliography{ref}

\end{document}

%% file: 1intro.tex
Machine learning (ML) techniques such as sequence-to-sequence models and graph neural networks have transformed many areas, including chat services and recommender systems~\citep{achiam2023gpt,pal2020pinnersage,corso2024graph}. These advances are now extending into the sciences, where even human experts may struggle to fully grasp the complexities~\citep{boiko2023autonomous,merchant2023scaling}. Many scientific discovery efforts have focused on small molecules (\textit{e.g.}, ChEMBL~\citep{gaulton2012chembl}, PubChem~\citep{kim2019pubchem}, OGB-LSC~\citep{hu2021ogb}) or inorganic crystals (\textit{e.g.}, Materials Project~\citep{jain2013commentary}, GNoME~\citep{merchant2023scaling}). However, limited attention has been directed toward \textit{polymers}, which are promising candidates for environmentally friendly and sustainable materials. For example, polymeric membranes are central to separation technologies that can greatly reduce the energy, carbon, and water intensity of many traditional thermally driven separation processes~\citep{sholl2016seven}.

Polymer discovery has been hindered by the absence of large-scale, open-source datasets~\citep{otsuka2011polyinfo}. Polymer data generation experimentally is slow because polymer synthesis involves complex polymerization reactions~\citep{kiesewetter2010organocatalysis} and processing steps~\citep{BAIRD2003611}. For instance, for gas separation polymers, decades of experimental efforts between 1950 and 2018 produced only about 1,500 polymers with gas permeability measurements, where each major gas type has 400 to 800 reported values. Generative models can propose hypothetical polymer structures~\citep{ma2020pi1m}, but these candidates often lack experimentally validated properties, limiting their practical utility.

\begin{figure}
    \centering
    \includegraphics[width=0.45\linewidth]{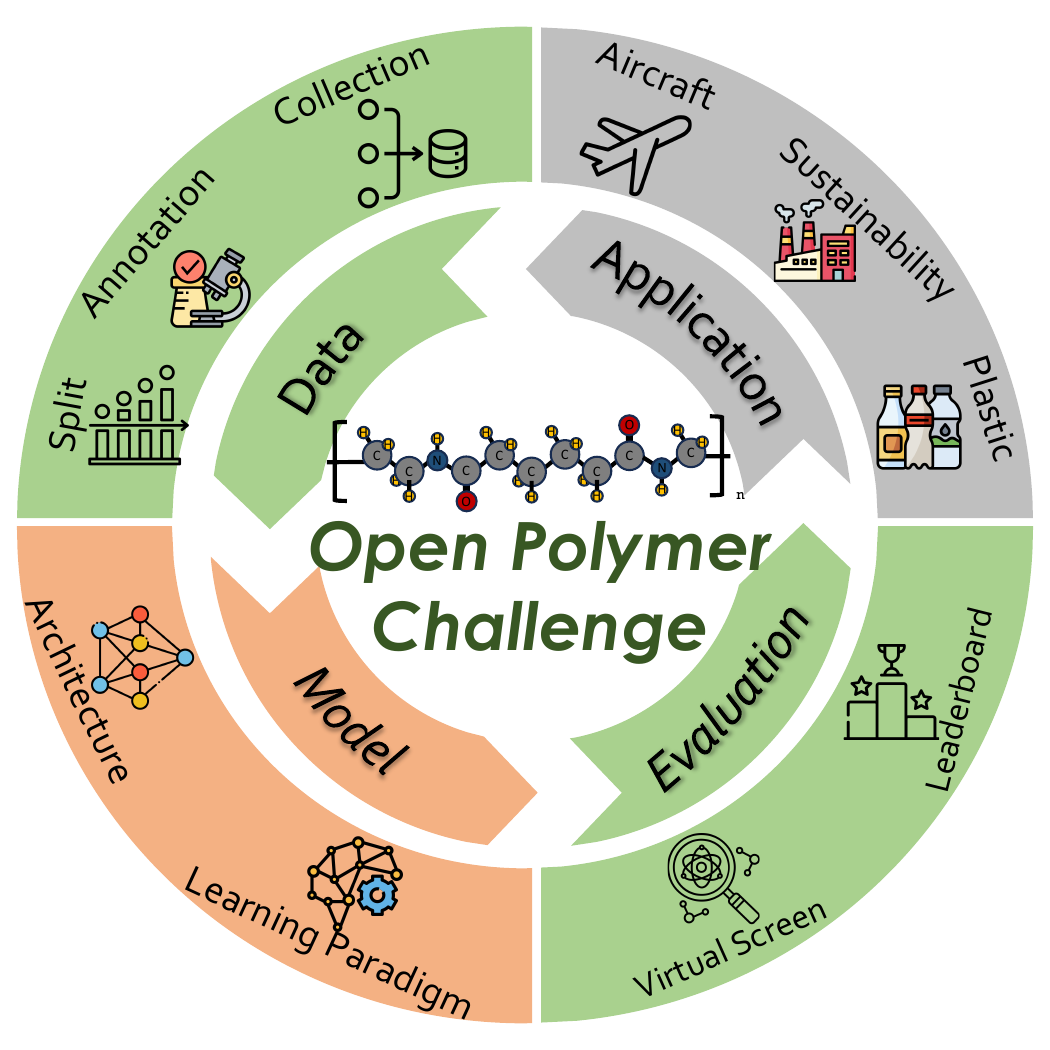}
    \caption{Overview of Open Polymer Challenge from four dimensions.}
    \label{fig:main}
\end{figure}

The scarcity of polymer datasets has slowed the progress of ML models designed specifically for polymers compared to other areas, such as small molecules and proteins. Motivated by this gap, we launched the community-driven Open Polymer Challenge (OPC), as presented in~\cref{fig:main}. We prepared polymer data and annotations and set up evaluation metrics to call for new solutions and models for polymer applications. There are five molecular dynamics (MD)-simulated polymer properties (\cref{tab:property-stats}): thermal conductivity (\tc), radius of gyration (\rg), density (\density), fractional free volume (\ffv), and glass transition temperature (\tg). To the best of our knowledge, this is the first ML competition dedicated to polymer informatics in both the ML and materials science communities. As summarized in \cref{table:related}, the OPC differs from existing competitions in biology, cheminformatics, and materials science, as it introduces new data objects and challenging multi-task prediction problems.

The competition revealed several important lessons about data preparation and property generation. Early in the challenge, a portion of previously generated polymer data that had once been publicly visible was unintentionally reused by participants. This happened even though the data had been removed from GitHub before the competition, but it had already been re-circulated by third parties. After detecting the overlap through SMILES matching on the leaderboard, we released that subset openly and created new polymer structures with new MD based property labels for both the public and private leaderboards.
These updates also exposed issues that often arise when simulations come from different groups, where we detail in~\cref{sec:pipeline}.

Although these challenges required careful handling, the competition still succeeded in producing reliable benchmarks and meaningful scientific insights. The competition ran on Kaggle from June 16, 2025 to September 15, 2025, spanning three months. It attracted more than 10,000 registrations, 2,600 participants, 2,000 teams, and 50,000 submissions from over 100 countries. The five prize-winning teams represented a wide range of backgrounds, including undergraduate and graduate students, as well as software engineers with or without prior chemistry experience. Analysis of their solutions shows that high-performing approaches relied on careful domain-specific data curation, fingerprint- and descriptor-based feature engineering, and strong but relatively simple models, such as gradient-boosted trees with rigorous cross-validation. These strategies yielded accurate predictions for $T_g$, density, FFV, and $R_g$ under small, noisy, and imbalanced data. The resulting best practices and creative ideas, including physics-informed features and structure-aware polymer augmentations, provide concrete guidance for future work in polymer property prediction and molecular AI. We discuss the new ML models and methods in~\cref{sec:result}.

\input{tables/table-related}

%% file: tables/table-related.tex
\begin{table}[!t]
\caption{Comparison between the proposed challenge and existing machine learning competitions (their latest version) on chemistry and material data.}
\centering
\begin{adjustbox}{width=0.98\textwidth}
\begin{tabular}{llllll}
\toprule
\textbf{Challenge} & \textbf{Venue} & \textbf{Domain} & \textbf{Data object} & \textbf{Task} & \textbf{Annotation method} \\
\midrule
\multirow{2}{*}{CellSeg} & NeurIPS & (1) Biomedicine & Cell & Cell & \multirow{2}{*}{Manual} \\
 & 2022 & (2) Cell & Image & segmentation & \\
\midrule
Single-cell & NeurIPS & (1) Drug & Cell \& & Differential gene & \multirow{2}{*}{Wet-lab} \\
perturbation & 2021--2023 & (2) Cell & Compound pair & expression value & \\
\midrule
OGB-LSC & KDD 2021 & (1) Quantum & Molecule & HOMO-LUMO & Density \\
(PCQM4Mv2) & NeurIPS 2022 & (2) Chemistry & Graph & energy gap & functional theory \\
\midrule
Open Catalyst & NeurIPS & (1) Catalyst & Adsorbate & Adsorption & Density \\
Challenge & 2021--2023 & (2) Materials & Catalyst surface & energy & functional theory \\
\midrule
BELKA & NeurIPS & (1) Drug & Small & Protein & DNA-encoded  \\
challenge & 2024 & (2) Biology & Molecule & binding affinity & library technology  \\
\midrule
\multicolumn{6}{c}{\textbf{Open Polymer Challenge:}} \\
\midrule
Open Polymer & Kaggle & Polymer & Polymer & Thermal, structural, and & \multirow{2}{*}{Molecular dynamics} \\
Challenge & \& NeurIPS 2025 & Materials & Structure & thermophysical properties & \\
\bottomrule
\end{tabular}
\end{adjustbox}
\label{table:related}
\end{table}

%% file: 2overview.tex
\input{tables/table_smiles}
\input{tables/table-property}

\paragraph{Problem Definition}

A polymer $X$ can be represented in two standard formats: (1) a SMILES string, which encodes the molecular structure as a sequence, and (2) a graph, where atoms are nodes and bonds are edges with defined chemical attributes. These two formats are chemically consistent and describe the same polymer structure. In this competition, we released the SMILES strings as polymer identifiers. We used five material properties: thermal conductivity (\tc), radius of gyration (\rg), mass density (\density), fractional free volume (\ffv), and glass transition temperature (\tg). All properties are obtained through MD simulations.

This challenge curated datasets containing 11,475 unique polymers, among which 9,625 have at least one labeled property. The distributions of structures and properties across the train, public leaderboard, and private leaderboard splits are given in~\cref{tab:smiles-counts,tab:property-stats}, together with their units and simulation time. Polymers were organized into three subsets: the train set used for model development, the public leaderboard set, and the private leaderboard set. Only the train set was released on Kaggle; both leaderboard sets remain hidden for evaluation.

\paragraph{Competition Metrics}
We evaluate the models using a weighted Mean Absolute Error (wMAE) across all properties, defined as
\begin{equation} \label{eq:wmae}
\text{wMAE} = \frac{1}{|\mathcal{X}|} \sum_{X \in \mathcal{X}} \sum_{i \in \mathcal{I}(X)} w_i \cdot | \hat{y}_i(X) - y_i(X) |,
\end{equation}
where $\mathcal{X}$ is the set of polymers being evaluated, and $\mathcal{I}(X)$ is the set of property types available for polymer $X$. \textbf{}The terms $\hat{y}_i(X)$ and $y_i(X)$ are the predicted and true values of the $i$-th property.

To balance contributions from properties with different scales and varying amounts of available data, we use a reweighting factor
\begin{equation} \label{eq:weight}
w_i = \left(\frac{1}{r_i}\right) \cdot
\left(
\frac{K \cdot \sqrt{1/n_i}}
{\sum_{j=1}^{K} \sqrt{1/n_j}}
\right),
\end{equation}
where $n_i$ is the number of available labels for the $i$-th property, and $r_i = \max(\mathcal{Y}_i) - \min(\mathcal{Y}_i)$ is its value range estimated from the training data. $K$ is the total number of prediction tasks.

This weighting scheme has two goals: (1) normalize for scale so that properties with large numerical ranges do not dominate the loss; (2) adjust for label imbalance through inverse square-root scaling so that rare properties receive appropriate weight.

%% file: tables/table_smiles.tex
\begin{table}[t]
\centering
\caption{Unique polymer structures in the Open Polymer Challenge. ``Total'' counts all SMILES in each dataset, while ``With any label'' counts only polymers containing at least one property label.}
\begin{adjustbox}{width=0.8\textwidth}
\begin{tabular}{lcc}
\toprule
Dataset & Total Unique SMILES & Unique SMILES With Any Label \\
\midrule
Train & 7,973 & 7,973 \\
Public Leaderboard & 295 & 295 \\
Private Leaderboard & 3,207 & 1,354 \\
\bottomrule
\end{tabular}
\end{adjustbox}
\label{tab:smiles-counts}
\end{table}

%% file: tables/table-property.tex
\begin{table}[t]
\centering
\caption{Polymer properties in the open polymer challenge. \ffv\ is fractional free volume, \tg\ is glass transition temperature, \tc\ is thermal conductivity, \rg\ is radius of gyration, and \density\ is mass density. CPU hours per sample correspond to the simulation cost for each property. Statistics include minimum, median, and maximum values for the train, public, and private datasets.}
\begin{adjustbox}{width=0.98\textwidth}
\begin{tabular}{llcccccc}
\toprule
Dataset & Property & Min & Median & Max & Count & Unit & CPU hrs/sample \\
\midrule

\multirow{5}{*}{Train}
& \tg & -148.030 & 74.040 & 472.250 & 511 & $^\circ$C & 2880 \\
& \ffv & 0.227 & 0.364 & 0.777 & 7030 & - & 864 \\
& \tc & 0.047 & 0.236 & 0.524 & 737 & W$\cdot$m$^{-1}\cdot$K$^{-1}$ & 1440 \\
& \rg & 9.728 & 15.052 & 34.673 & 614 & \AA & 432 \\
& \density & 0.749 & 0.948 & 1.841 & 613 & g$\cdot$cm$^{-3}$ & 432 \\
\midrule

\multirow{5}{*}{Public Leaderboard}
& \tg & -138.802 & 97.764 & 359.084 & 95 & $^\circ$C & 2880 \\
& \ffv & 0.315 & 0.353 & 0.656 & 86 & - & 864 \\
& \tc & 0.077 & 0.244 & 0.678 & 239 & W$\cdot$m$^{-1}\cdot$K$^{-1}$ & 1440 \\
& \rg & 11.208 & 17.436 & 32.301 & 54 & \AA & 432 \\
& \density & 0.747 & 1.057 & 1.360 & 244 & g$\cdot$cm$^{-3}$ & 432 \\
\midrule

\multirow{5}{*}{Private Leaderboard}
& \tg & -86.650 & 169.670 & 484.690 & 166 & $^\circ$C & 2880 \\
& \ffv & 0.306 & 0.351 & 0.454 & 137 & - & 864 \\
& \tc & 0.024 & 0.239 & 0.954 & 1165 & W$\cdot$m$^{-1}\cdot$K$^{-1}$ & 1440 \\
& \rg & 9.445 & 18.511 & 35.530 & 1056 & \AA & 432 \\
& \density & 0.135 & 1.081 & 1.733 & 1282 & g$\cdot$cm$^{-3}$ & 432 \\
\bottomrule
\end{tabular}
\end{adjustbox}
\label{tab:property-stats}
\end{table}


%% file: 3pipeline.tex
\begin{figure}[t]
    \centering
    \includegraphics[width=0.92\linewidth]{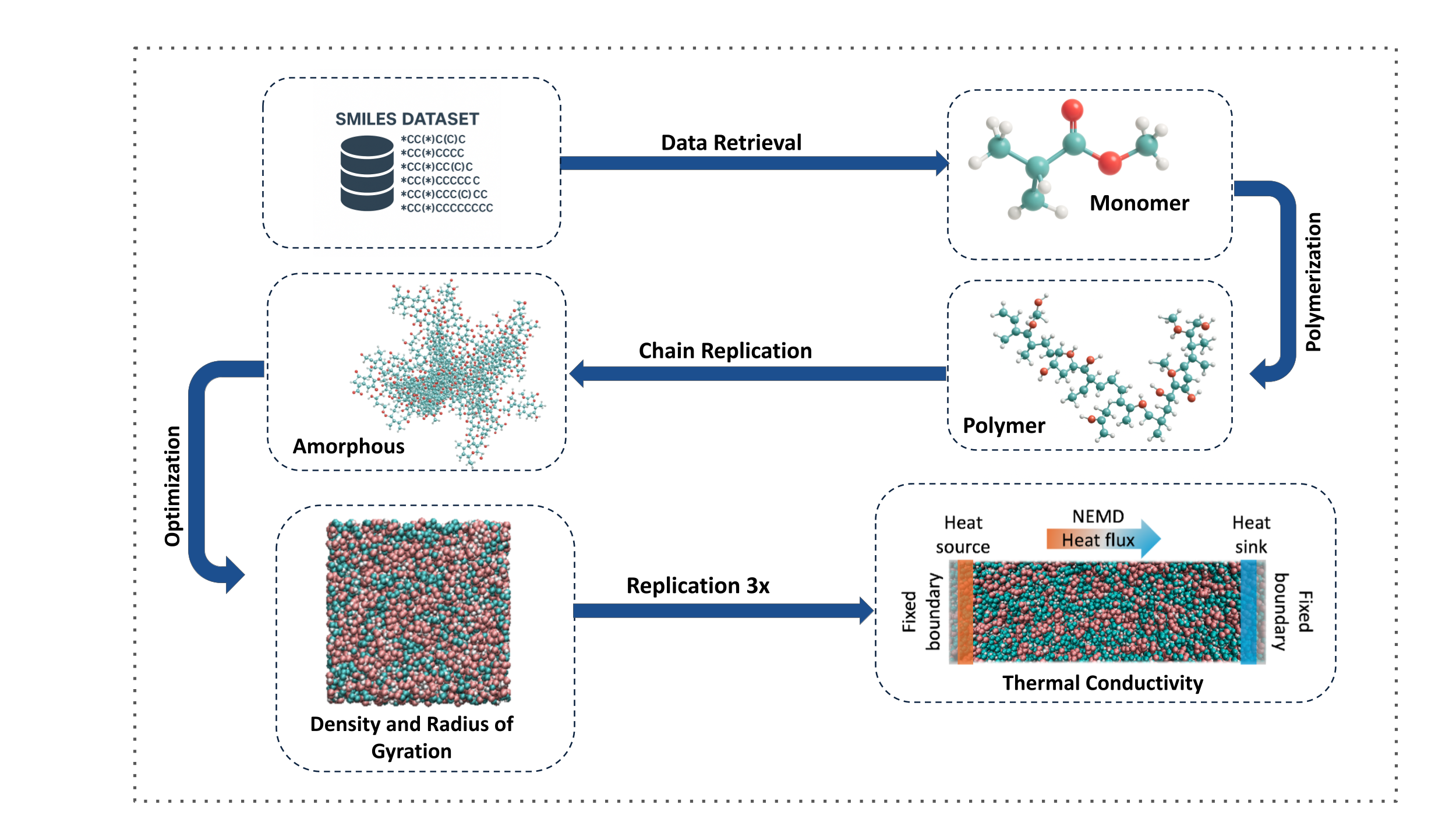}
    \caption{An example of a high-throughput MD workflow for polymer property calculations. The process includes SMILES parsing, monomer-to-polymer chain construction, amorphous packing, multi-stage equilibration, and computation of density, radius of gyration, and thermal conductivity.}
    \label{fig:workflow}
\end{figure}

Due to the detection of data leakage early in the competition, we generated new evaluation sets for all five properties during the challenge. To prepare new polymer structures, we used three sources. First, we sampled 100 structures from the PI1M database containing one-million virtual polymers ~\citep{ma2020pi1m} that had no overlap with the competition data. Second, we used the PI1M generative model~\citep{ma2020pi1m} to create 2,000 previously unseen structures. Third, we added 900 experimentally validated polymers that lacked competition labels as a backup pool. In total, we prepared 3,000 candidate structures, forming a superset from which the final public and private test sets were drawn. This avoided repeated updates and prevented teams from re-running inference after each correction. Two research groups simulated the polymers in parallel, one for \tc, \density, and \rg, and the other for \tg and \ffv.

There were two rounds of data updates. The first update was public and caused shifts in the public leaderboard after the new data were released. During this step, we identified RDKit version conflicts: about 20 of the 3,000 SMILES strings were valid in the older RDKit versions but invalid in newer ones. We also found a unit mismatch (K <-> \(^{\circ}\mathrm{C}\)) in the new \tg values. Both issues were corrected. The second update affected only the private leaderboard. In this stage, we observed a shift in the mean \tg between the training data and the private test set. A detailed check (see~\cref{subsec:property-generation}) showed that the shift is not from incorrect units.

These observations highlight three lessons. First, even brief public release of intermediate data can propagate widely and create leakage. Second, consistency across simulation batches requires careful checks of units, fitting methods, and property definitions. Third, dynamic data generation during a live competition introduces the risk of distribution shifts that cannot be curated or stabilized in advance. 
Next, we detail the methods and pipelines used in data generation.

\subsection{Amorphous Polymer Structure Generation}

To construct the amorphous polymer systems, we employed a high-throughput MD workflow, named ADEPT: Automated molecular Dynamics Engine for Polymer simulaTions, consisting of monomer-to-polymer structure generation, amorphous cell packing, and multi-stage equilibration (available at \url{https://github.com/sobinalosious/ADEPT}). Figure~\ref{fig:workflow} illustrates the overall MD workflow used for polymer generation, amorphous packing, and property calculations. Polymer monomers were represented using simplified molecular input line entry system (SMILES) strings~\cite{weininger1988smiles}. A Python-based pipeline built on PySIMM~\cite{fortunato2017pysimm} was used to generate polymer chains from these monomers. Linear chains were grown using a random-walk polymerization algorithm, followed by methyl (–CH\textsubscript{3}) termination at both chain ends. For the density, radius of gyration and TC calculations, the General AMBER Force Field 2 (GAFF2)~\cite{vassetti2019assessment} was applied to assign force field parameters, and Gasteiger charges were used for electrostatics. Each chain contained approximately 600 atoms, and six chains were placed randomly in a periodic simulation box at low initial density to form an amorphous cell. Input files compatible with LAMMPS~\cite{plimpton1995fast} were automatically generated by the pipeline for subsequent MD simulations. 

The systems were then equilibrated through two successive stages: initial relaxation and annealing. During initial relaxation, electrostatic interactions were disabled and Lennard--Jones (LJ) interactions were truncated at 0.3~nm to alleviate large forces from close contacts. The system was equilibrated in the NPT ensemble at 100~K for 2~ps with a 0.1~fs timestep, followed by heating to 1000~K over 1~ns in the NVT ensemble. Next, NPT simulations were conducted at 1000~K and 0.1~atm for 50~ps and subsequently at pressures ramped from 0.1~atm to 500~atm for 1~ns using a 1~fs timestep. SHAKE constraints~\cite{ryckaert1977numerical} were applied to maintain covalent bond lengths.

In the annealing stage, electrostatics were re-enabled using the particle--particle--particle--mesh (PPPM) Ewald summation method~\cite{hockney2021computer}, and the LJ cutoff was increased to 0.8~nm. The system was equilibrated in an NPT ensemble at 1000~K and 1~atm for 2~ps using a 0.1~fs timestep, then cooled to 300~K at a rate of 140~K/ns with SHAKE enabled. Finally, an 8~ns NPT simulation was performed at 300~K and 1~atm with a 1~fs timestep to obtain a well-equilibrated amorphous configuration suitable for subsequent density, radius of gyration, and thermal conductivity calculations (see Fig.~\ref{fig:workflow}). The simulation box was allowed to relax but constrained to remain cubic.

During the simulation, data consistency was maintained through a fully automated Python–LAMMPS workflow that standardized system generation, force field assignment, equilibration, and analysis across all polymers. Within each of the two data generation groups, identical simulation parameters and force fields were applied to each system, and input scripts were auto-generated to avoid manual errors. Post-processing was performed using uniform Python routines with checks for trajectory length, thermodynamic stability, and file integrity. Repeated property calculations followed the same analysis scripts to ensure reproducibility and comparability across the dataset.

The generation of the amorphous polymer system for \ffv and \tg followed a different workflow. Material Studio was used to create polymer chains with $\sim$2000 atoms and 20 identical chains were packed into a simulation box to the initial amorphous polymer system.\cite{tao2021benchmarking} The initial system was optimized with polymer consistent force field (PCFF)~\cite{pcff} under a 3D periodic boundary condition using a 21-step equilibration protocol~\cite{equil21step}. The LAMMPS simulation package was used for all simulations.
\subsection{Property Generation}\label{subsec:property-generation}

\paragraph{Density}
The density of each polymer system was calculated from NPT simulations conducted at 300~K and 1~atm following the annealing stage. After equilibration, the simulation was continued for a production run, and instantaneous system density was recorded at regular intervals using LAMMPS' built-in \texttt{variable density} command. The final reported density corresponds to the time-averaged value over this production window.

Density was evaluated only after long NPT equilibration runs to ensure that each polymer system reached a stable volume and energy. For every trajectory, we identified a stable sampling window and restricted the calculation to the final 3~ns, where volume fluctuations were minimal. Density values were reported only for systems in which the complete simulation finished without interruption or numerical failure.

\paragraph{Radius of Gyration (\rg)}
The radius of gyration (\rg) of individual polymer chains was computed during the same NPT production run at 300~K and 1~atm. Per-molecule chunking was performed using \texttt{compute chunk/atom molecule}, and \rg values were calculated for each chunk using \texttt{compute gyration/chunk}. Time-averaging was carried out using \texttt{fix ave/time} over the entire production window. Reported \rg values correspond to the temporal average and are summarized as mean~$\pm$~standard deviation over all chains. The radius of gyration is given by
\begin{equation}
R_g = \sqrt{\frac{1}{N} \sum_{i=1}^{N} \left\| \mathbf{r}_i - \mathbf{r}_{\mathrm{cm}} \right\|^2 },
\end{equation}
where $N$ is the number of atoms in the chain, $\mathbf{r}_i$ are the atomic positions, and $\mathbf{r}_{\mathrm{cm}}$ is the chain’s center of mass.
For the radius of gyration, polymer chains were consistently identified using molecule IDs so that atoms were never mixed between different chains. $R_\mathrm{g}$ values were extracted only from fully equilibrated amorphous cells, avoiding intermediate configurations that were still stretched or compressed. To reduce the impact of thermal noise, instantaneous $R_\mathrm{g}$ values were averaged over long time windows, providing smooth and representative chain statistics.

\paragraph{Thermal Conductivity (\tc)}
Thermal conductivity was computed along the $x$-direction using a non-equilibrium MD (NEMD) approach~\cite{MA2022100850}. After equilibration, the simulation cell was replicated $3 \times 1 \times 1$ along $x$. Narrow source and sink slabs were defined symmetrically around the box center, and atoms in these slabs were thermostatted using Langevin dynamics at 320~K (source) and 280~K (sink), respectively. The central region evolved under NVE dynamics. A steady-state temperature profile was established by running the simulation for $2 \times 10^7$ steps with a 0.25~fs timestep (5~ns total). Temperature profiles were collected by binning the system along $x$ using the commands \texttt{compute chunk/atom bin/1d x ...} and \texttt{fix ave/chunk}. The temperature gradient $\mathrm{d}T/\mathrm{d}x$ was obtained by fitting only the central linear region. Heat flux $J_q$ was computed from the cumulative energy added to the hot slab and removed from the cold slab, normalized by cross-sectional area. Thermal conductivity $\kappa_x$ was then determined via Fourier’s law,
\begin{equation}
\kappa_x = -\,\frac{J_q}{\mathrm{d}T/\mathrm{d}x}.
\end{equation}
Multiple non-overlapping time segments were analyzed independently, and the mean and standard deviation of $\kappa_x$ were reported.

Thermal conductivity calculations were accepted only after establishing a clear steady state in the non-equilibrium simulations. Heat flux and temperature profiles were monitored, and $\kappa$ was computed only once a linear, stable temperature gradient was observed. Energy added to the hot region and removed from the cold region was compared to ensure bidirectional balance and to rule out numerical drift. The steady-state portion of each trajectory was further divided into several independent time blocks; $\kappa$ was calculated for each block, and the reported value corresponds to the block-averaged mean. The temperature-gradient fit was restricted to the central conduction region, excluding noisy bins near the thermostatted zones. Any unphysical negative $\kappa$ values were discarded during post-processing.

\paragraph{Fractional Free Volume (\ffv)}
For \ffv calculations, NPT simulations at 300 K and 1 atm were performed for 5,000,000 steps using a timestep of 1 fs, totaling 5 ns. Certain simulations were unstable at 1 fs, for such systems smaller timesteps of 0.75, 0.5, and 0.25 fs were attempted. Polymer systems that were not stable at 0.25 fs were ignored. A total of 10 structures from the last 1,000,000 steps of the simulation was used to evaluate \ffv using PoreBlazer 4.0~\cite{poreblazer}. 

PoreBlazer 4.0 can evaluate \ffv of two types: total and network-accessible (percolated). For each of these types, the \ffv is occupiable with He and N$_2$, and geometric are evaluated by the software. The reported \ffv data was the total geometric \ffv (ratio of total \verb{V_G_A^3{ and \verb{V_A^3{ in the \verb{summary.dat{ output file generated by PoreBlazer 4.0)~\cite{tao2023machine}. Initially, the default total N$_2$ probe \ffv was used to determine the \ffv. This resulted in several near-zero \ffv that were outside the training range. 
This issue was quickly identified, and accurate geometric \ffv were added. The uploaded data had no data inconsistency; All data were generated using Poreblazer 4.0 and geometric \ffv for each polymer was determined.

Since FFV are measured using standard software, no data distribution shift are expected. Observations of better prediction accuracy with $\text{FFV}^2$ was again a mere coincidence. Since, $0<\text{FFV}<1$, squaring FFV, reduces its magnitude. Such strategies might improve model performance if output FFV is known to be lower, however it will perform much worse when output FFVs have higher mean. 

\paragraph{Glass Transition Temperature (\tg)}
\begin{figure}[tb]
    \centering
    \includegraphics[width=0.92\linewidth]{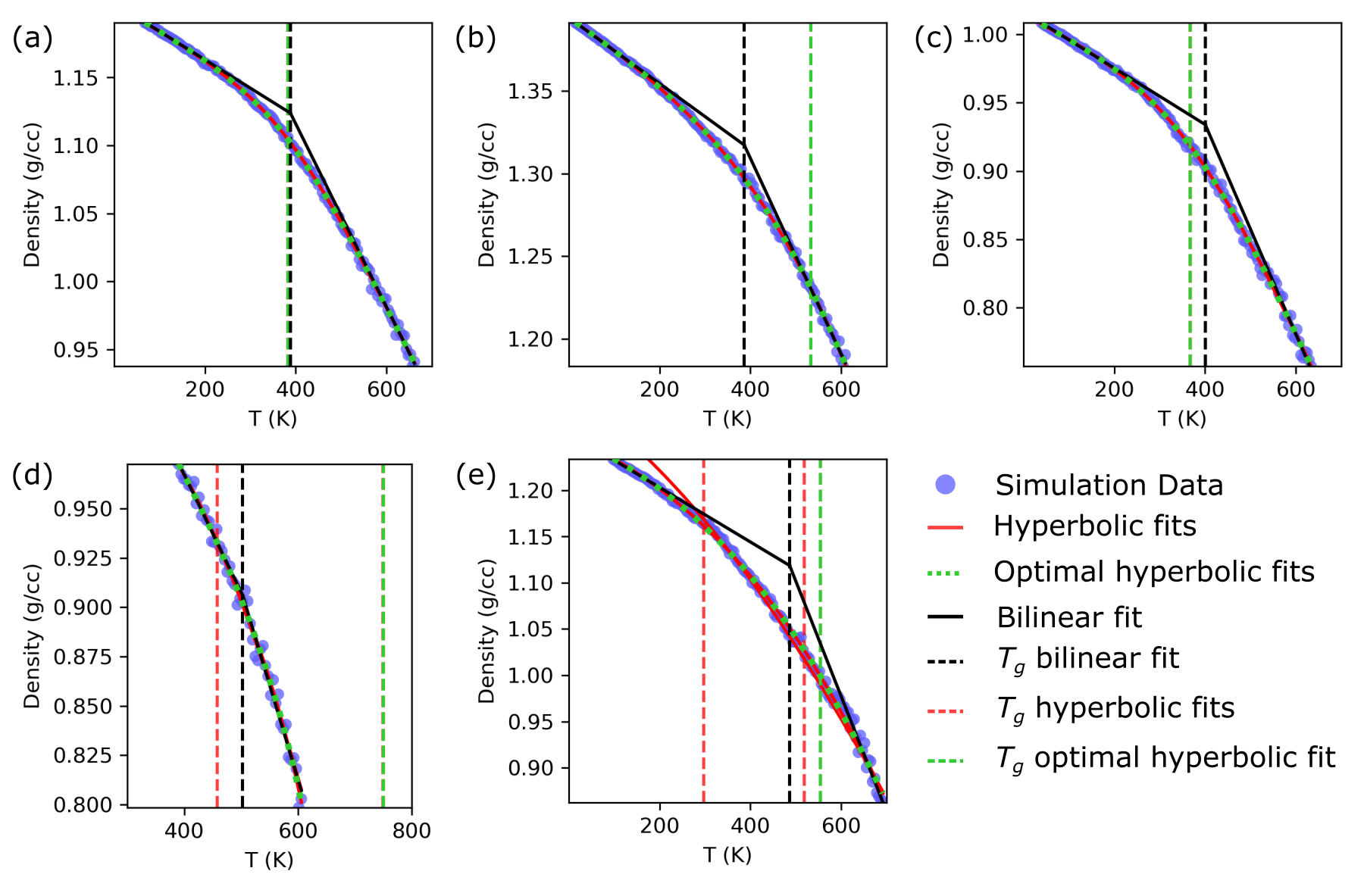}
    \caption{Glass transition temperature (\tg) from bi-linear and hyperbolic fits. (a) Early test cases where bi-linear and hyperbolic fits produced identical results. Later verifications showed (c, d) good hyperbolic fits can cause significant deviations in \tg, and (d, e) produce wide range of \tg that is dependent on the constraint bounds on fitting parameters. (d) Best hyperbolic fit can be high.}
    \label{fig:Tg}
\end{figure}

The system was initially equilibrated at 650 K (time constant 0.1 ps) and 1 bar (time constant 1 ps, drag 0.2) for 1 ns. Thereafter, at the same pressure, the system was cooled to 150 K over 10 ns. The density of the polymer system was plotted as a function of the system temperature. 

The \tg values from density-temperature plots can be evaluated using two popular methods: (i) bi-linear fit, where two linear asymptotes are evaluated, and their intersection provides the \tg, (ii) hyperbolic fit, where the focus of the hyperbola yields the \tg~\cite{tghyperbolic}. For hyperbolic fits, all data are fitted to Eq.~\ref{eq:tg_hyper}, where $(\rho_0,T_g)$ is the hyperbola center, and slopes $-a$ and $-(a+b)$ are the asymptotes at low and high temperature regimes, where $a,b>0$. As $c\to -\infty$, the hyperbola approaches a piecewise linear function (similar to a bilinear fit).c

\begin{equation}
\label{eq:tg_hyper}
    \rho(T) = \rho_0 -\left(a-\frac{b}{2}\right)(T-T_g) -b\sqrt{\frac{(T-T_g)^2}{4} + e^c} \quad \quad a>0, b>0
\end{equation}

There are inherent issues with the bi-linear fit method, as one has to choose an arbitrary range over which the linear fit is performed. The optimal range for linear fits can differ based on the data. Therefore, hyperbolic fits were used to perform a high-throughput evaluation of \tg. Since hyperbolic fits are used over the entire temperature range, only limited constraints are necessary (Eq.~\ref{eq:tg_hyper}). While generating the test data for the public leaderboard, consistency tests were performed for several polymers, which yielded good results (Fig.~\ref{fig:Tg} (a)). With the initial consistency, the public leaderboard data were generated using hyperbolic fits. 
However, comprehensive checks after the Kaggle competition showed inconsistencies between \tg generated with bi-linear fit and hyperbolic fitting methods (Fig.~\ref{fig:Tg}~(b)-(e)). Many \tg values evaluated with hyperbolic fits are higher than the corresponding values from the bi-linear fit (Fig.~\ref{fig:Tg}(b)-(e)). Another issue was that \tg values from the hyperbolic fits were dependent on the constraints put on \tg. These issues caused some of the shifts in the mean \tg and were not associated with unit conversions. We note that polymers of the order of thousands were considered in the competition but total possibilities of polymers can be billions-and-billions. In such cases differences in data distribution is unavoidable.

We note that the data distributions were not associated with $^\circ$C to $^\circ$F conversions. The public leaderboard data associated with \tg had a mean and standard deviation of $\mu_{pub}=102.9$ and $\sigma_{pub}=103.7$, respectively. The private leaderboard data associated with $T_g$ had a mean and standard deviation of $\mu_{pri}=179.8$ and $\sigma_{pri}=134.9$, respectively. 
Let us consider a linear transformation $y=ax+b$ that converts public leaderboard data to private leaderboard. Then, 

\begin{equation}
\label{eq:tg_transform}
\begin{split}
    a = \frac{\sigma_{pri}}{\sigma_{pub}}=1.3 \\
    b = \mu_{pri}-a\mu_{pub}=46 
\end{split}
\end{equation}

For conversions from $^\circ$C to $^\circ$F, the expected constants are $a=1.8$ and $b=32$, which differ clearly from the observed values. 

%% file: 4result.tex
\subsection{Winner solutions}

While each of the top solutions introduced unique innovations, they all converged on similar strategies, including additional data collection and cleaning, data augmentation, \tg label shift correction, fingerprint-based feature selection, and cross-validation. These shared solution features enabled participants to create models that achieved high predictive performance in this competition.

To remedy the limitation of data scarcity in the competition, the majority of the top solutions collected additional polymer data to expand the labeled training sets. Aside from the original training and supplementary data, most participants chose to further incorporate additional data sources of \tg~\cite{choi2024automated, borredon2023characterising, otsuka2011polyinfo} and \density~\cite{hu2021chemprops, hayashi2022radonpy}. Additionally, the 1st place solution chose to perform fast MD simulations with neural networks to calculate \density, \ffv, and \rg for 1116 polymers from PI1M~\cite{ma2020pi1m}. These additional data sources aid model training by expanding coverage in the chemical space and reducing overfitting.

Most teams cleaned the data by performing canonicalization and kekulization to clean and deduplicate the training data. However, 3 of the top-10 solutions chose to use non-canonical forms of each polymer SMILES to generate new training data. Some teams used other forms of data augmentation as well, such as the 3rd place representing polymers as multi-monomer chains explicitly or the 4th place substituting functional groups with similar groups. The 1st place chose to perform fairly extensive data filtering, first by filtering training data whose predictions from an ensemble model exceed an error threshold and then by removing out-of-distribution samples, specifically from the \tc data. The team found that removing polymers with \tc values over 0.402 led to better results.

The final data consideration that most teams addressed was the shift in the \tg labels between the training and leaderboard test data. Teams chose to scale the data by manually selecting a constant bias term or train a linear regressor to find the optimal weight and bias terms. The fourth-place team found that by using a quantile regression target, which guides the model to learn from values higher or lower in the label distribution (for example, using an objective percentile $\alpha = 0.85$), they could improve predictive accuracy for \tg and \rg. These data processing strategies led to the success of these top solutions.

\paragraph{Feature Selection}
The features often selected by the top solutions were fingerprints and molecular descriptors. All solutions in the top-10 used Morgan fingerprints~\cite{rogers2010extended}. RDKit descriptors~\cite{rdkit}, MACCS~\cite{maccs}, Atom Pair~\cite{atompair_fingerprint}, Topological Torsion~\cite{topological_torsion}, Mordred descriptors~\cite{modred}, and ECFP~\cite{rogers2010extended} were also used, listed in descending order of use in feature selection. The top-3 solutions used graph features, although only the 3rd place team used a graph neural network to represent the polymer. The 5th place used \ffv and \tc prediction values as features to predict \tg, \rg, and \density, the rationale being that \ffv and \tc data were more abundant and were closely linked to the other properties according to their observations. The 1st place was unique among the top solutions for utilizing uncommon features, including shape-based descriptors, Gasteiger charge, element composition, bond type ratios, and polyBERT embeddings. The 1st and the 3rd places explicitly mentioned filtering features by those that were most predictive. Despite the advances in sequence processing and graph learning, SMILES embeddings and graphs were not commonly used, each only appearing in one of the top-10 solutions.

\paragraph{Model and Training}
Modeling strategies were surprisingly diverse. While most of the top solutions used tree-based models such as XGBoost~\cite{XGBOOST}, LightGBM~\cite{ke2017lightgbm}, CatBoost~\cite{catboost}, some used other models. The 1st place used an ensemble of polyBERT~\cite{kuenneth2023polybert}, AutoGluon~\cite{autogluon}, and Uni-Mol~\cite{alcaide2024unimoldockingv2realistic}, the 3rd place used a GAT, and the 5th place implemented a physics-based mathematical model with their tree ensemble. For model training, all solutions used 5-fold cross-validation. Notably, the 1st place performed BERT pretraining on self-supervised synthetic labels before finetuning.

\subsection{Best Practice in Polymer Property Prediction}

A key challenge in the competition was the inherently small scale, imbalanced, and noisy nature of the available polymer property datasets. With only a few thousand samples, the top 200 solutions repeatedly demonstrated that domain-specific data curation, the use of simple, well-tuned models combined with complementary signals or restrained ensembling, and disciplined learning strategies were key to achieving robust generalization.

\paragraph{Mitigating small-data constraints.}
A recurring strength among high-performing solutions was their careful adaptation to the limited and noisy nature of polymer data. Efficient techniques included canonicalization (mapping each SMILES to a unique standard representation) and kekulization (explicitly assigning single and double bonds to aromatic systems), which generated chemically valid sequences without altering molecular identity. In contrast, aggressive augmentation such as random stereoisomer or tautomer enumeration (listing all valid stereochemical variants) frequently led to overfitting, with the 3rd place team explicitly reporting its ineffectiveness for GNNs. Empirically, top solutions avoided unvalidated augmentation, instead focusing on representation-level diversity grounded in chemical invariance and test-time robustness.

\begin{figure}[htb]
    \centering
    \includegraphics[width=0.7\linewidth]{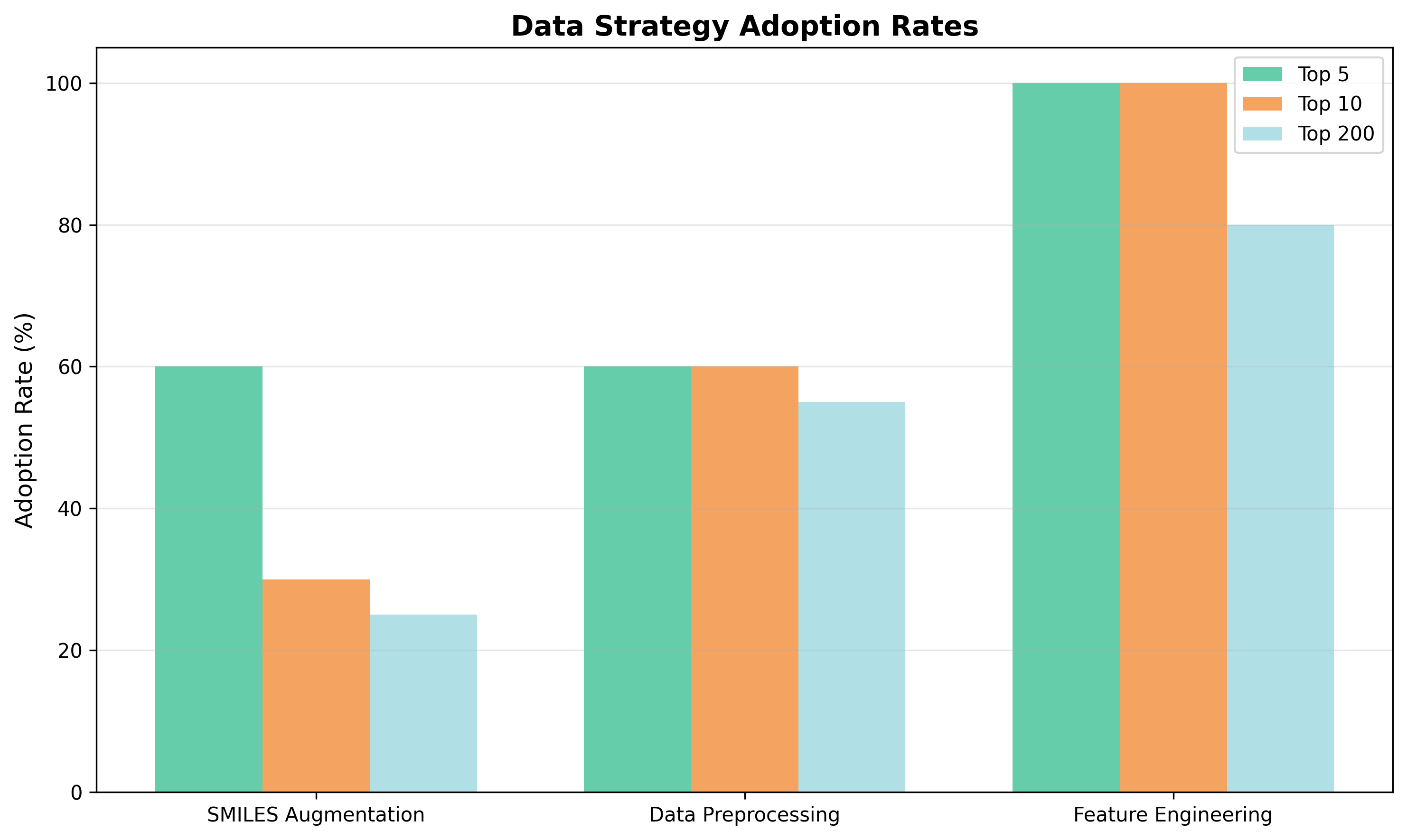}
    \caption{Choices of data strategies among Top 5, Top 10, and Top 200 leaderboard teams. Feature engineering (e.g., custom descriptors, fingerprints) was universally adopted in the top tiers, while use of SMILES augmentation showed decreasing adoption beyond Top 10.}
    \label{fig:data_aug}
\end{figure}

To expand structural expressiveness in small-data regimes, top-performing teams extensively constructed feature pools, combining classical descriptors (e.g., RDKit~\cite{rdkit}, Mordred~\cite{modred}) with diverse fingerprints (e.g., Morgan/ECFP~\cite{rogers2010extended}, MACCS~\cite{maccs}, Atom Pair~\cite{atompair_fingerprint}, Topological Torsion~\cite{topological_torsion}), and in some cases, 3D geometric or MD-derived attributes. Feature counts often exceeded thousands, necessitating rigorous feature selection. The choices of different data strategies among top teams are illustrated in \cref{fig:data_aug}.

A common pipeline involved training preliminary tree-based models (e.g., XGBoost~\cite{XGBOOST}, LightGBM~\cite{ke2017lightgbm}, CatBoost~\cite{catboost}) to compute per-family importance scores and retain only a top-$N$ subset before fitting the final models. Automated hyperparameter search (e.g., Optuna~\cite{optuna_2019}) was also used to tune the number of retained features, drop entire feature categories, or adjust fingerprint dimensionality. Notably, the 10th place team implemented a robust multi-stage strategy: statistical gating, keeping descriptors only if they passed correlation or mutual-information tests with the target, followed by collinearity pruning to remove redundant signals. Such workflows preserved chemical signal while improving robustness under sparse supervision. 

\begin{figure}[htb]
    \centering
    \includegraphics[width=0.85\linewidth]{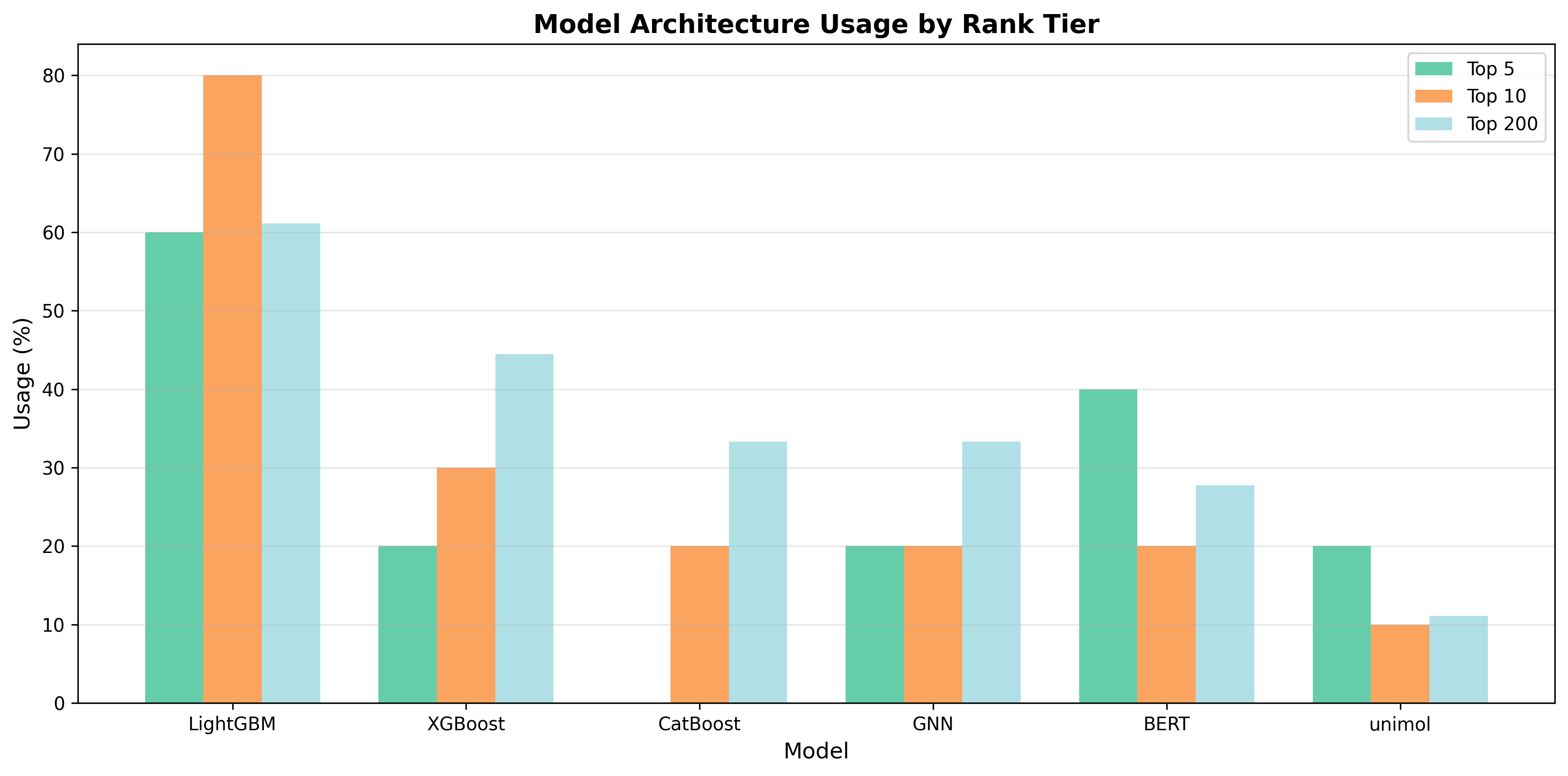}
    \caption{Model architecture usage among Top 5, Top 10, and Top 200 leaderboard teams. Tree-based models, especially LightGBM, remained dominant in top positions.}
    \label{fig:model_use}
\end{figure}

Modeling strategies under small-data conditions emphasized robustness over complexity. Tree-based methods (e.g., LightGBM~\cite{ke2017lightgbm}, XGBoost~\cite{XGBOOST}, CatBoost~\cite{catboost}, TabPFN~\cite{tabpfn}) consistently served as robust baselines, appearing in all top-5 solutions. Simple configurations with well-engineered features proved stronger than complex neural networks. The 14th-place team observed an inverse correlation between model complexity and performance. While several teams experimented with pretrained transformers (e.g., ChemBERTa~\cite{chemberta}, ModernBERT~\cite{modernbert}, CodeBERT~\cite{feng2020codebertpretrainedmodelprogramming}) or GNN methods (e.g., GATv2Conv~\cite{brody2022attentivegraphattentionnetworks}, GREA~\cite{liu2022graph}), these models required careful pretraining or ensembling to succeed. Most top teams employed lightweight ensembles, such as target-wise averaging of model outputs rather than deep fusion, to maintain stability. AutoGluon~\cite{autogluon} was also adopted in some solutions as an alternative to manually constructed model stacks for ensemble training under resource constraints. These modeling choices are quantitatively summarized in Figure~\ref{fig:model_use}, which highlights the prevalence of tree-based methods and the more selective use of neural architectures in top-ranking solutions.

\paragraph{Addressing data imbalance and distribution skew.}
The challenge’s multi-target setting introduced heterogeneity across properties: differing scales, label densities, and noise levels. To address this, top teams employed complementary strategies as follows:

Several teams leveraged external datasets from public resources to enrich under-sampled regions. Rather than direct concatenation, they performed rigorous distributional checks, including examining value ranges, SMILES consistency, and MAE gaps on overlapping entries. Retained datasets were harmonized via dataset-specific linear transformation, bin-wise resampling, or MAE-driven calibration. Simple regressions on overlapping samples were often sufficient to estimate optimal slope and intercept corrections. Once aligned, supplemental data boosted model performance.

Stratified K-fold validation with per-target quantile binning was widely adopted to ensure balanced folds and reliable error estimates~\cite{kfold}. Most teams trained separate single-task models for each target to isolate noise from underrepresented targets. This setup was often paired with out-of-fold predictions to construct meta-ensembles without leakage. To further reduce systematic bias, lightweight post-processing techniques—including isotonic regression, mean alignment, property-specific clipping, or fold-wise linear calibration—were employed, especially for tail-heavy properties where raw models tended to underpredict extremes.

For distribution shifts, participants applied target-wise normalization—for example, max-scaling, or applying heuristic offsets ($+20^\circ\mathrm{C}$ to $+40^\circ\mathrm{C}$) based on cross-validation feedback and leaderboard probing. Similar corrections were also used for the \density property and \rg, yielding more stable and calibrated predictions. For example, the 3rd-place team used fold-wise regression calibration, improving scores by 5.5\%. These findings highlighted the importance of error calibration, fold-aware correction, and diagnostic probing for robustness under real-world shifts.

\subsection{Creative Strategies and Open Challenges}

Beyond the dominant themes of data cleaning and ensemble modeling, several participants explored ideas that offered promising directions for advancing polymer informatics.

\textbf{Structure-aware augmentations grounded in polymer semantics.}
One idea was the recognition that monomer-level SMILES representations often lacked enough topological context to capture repeating patterns in real polymers. Some participants (the 3rd place and the 200th place) proposed SMILES level augmentation through polymer chain extension by connecting the polymerization points to generate plausible dimers or trimers. They suggested that chain extension could reflect the physical repeat structure of polymers rather than only their static composition, and that it could improve performance relative to isomer enumeration for GNN models. This approach aligned with the repetition augmentation introduced in a recent study on repetition invariant representations~\cite{grin}.

\textbf{Integrating physics priors through hybrid modeling.}
While most solutions relied on neural networks, select participants incorporated domain knowledge into both input features and prediction logic. The 5th place team blended physics-inspired descriptors with ML-based predictions through hierarchical modeling. \ffv and \tc were first predicted and then used as inputs for \density and \rg. Others manually constructed feature bundles that captured rigidity, hydrogen bonding, and packing heuristics, which are mechanistic cues often missed by learned representations. These methods showed that domain knowledge can improve performance even when models remain simple.

\textbf{Persistent challenge: data shift.}
Distributional robustness remained a major difficulty in polymer property prediction, especially when datasets were sourced from heterogeneous simulations or experimental protocols. Polymer annotations are sensitive to force field parameters and post-processing pipelines, and even small methodological differences can produce meaningful shifts that reduce model generalization. This issue was clearly seen during the competition. Several teams identified systematic distribution shift, especially in \tg, as the most influential factor that affected leaderboard outcomes. For example, the 1st place solution applied a calibrated offset proportional to the standard deviation ($T_g \leftarrow T_g + \sigma \times 0.5644$). These corrections, however, depended heavily on post-hoc public leaderboard probing and manual calibration rather than systematic diagnostics, showing the fragility of current pipelines when hidden shifts are present.

%% file: 5conclusion.tex
The Open Polymer Challenge created the first large-scale, community-driven benchmark for polymer property prediction and revealed both the promise and the current limitations of molecular AI in polymer science. Participants demonstrated that careful data curation, targeted augmentation, and strong but simple models can reach high accuracy even under small, noisy, and imbalanced datasets. The competition also exposed critical challenges in data generation, distribution shifts, and cross-group simulation consistency, emphasizing the need for transparent pipelines and standardized polymer datasets. Beyond the leaderboard results, the creative strategies and practical insights from hundreds of teams provide a foundation for future advances in polymer informatics. We hope that this benchmark, together with the released data, code, and analysis, will support sustained progress in molecular AI for materials discovery and motivate further collaboration between the ML and polymer science communities.